\documentclass[10pt,twocolumn,letterpaper]{article}

\usepackage{cvpr}
\usepackage{times}
\usepackage{epsfig}
\usepackage{graphicx}
\usepackage{amsmath}
\usepackage{amssymb}
\usepackage[font={small}]{caption}

\usepackage{url}            
\usepackage{booktabs}       
\usepackage{nicefrac}       
\usepackage{microtype}      
\usepackage[inline]{enumitem}
\usepackage{color}

\usepackage{footnote}
\usepackage{footmisc}

\usepackage{makecell}

\newcommand{\beas}{\begin{eqnarray*}}
	\newcommand{\eeas}{\end{eqnarray*}}
\newcommand{\bea}{\begin{eqnarray}}
\newcommand{\eea}{\end{eqnarray}}
\newcommand{\bes}{\begin{equation*}}
\newcommand{\ees}{\end{equation*}}
\newcommand{\be}{\begin{equation}}
\newcommand{\ee}{\end{equation}}

\newcommand{\norm}[1]{\left\lVert#1\right\rVert}

\newcommand{\alex}[1]{{\color{green}{Alex: #1}}}
\newcommand{\tamir}[1]{{\color{blue}{Tamir: #1}}}

\newcommand\blfootnote[1]{%
  \begingroup
  \renewcommand\thefootnote{}\footnote{#1}%
  \addtocounter{footnote}{-1}%
  \endgroup
}

\makeatletter
\usepackage{xspace}
\def\@onedot{\ifx\@let@token.\else.\null\fi\xspace}
\DeclareRobustCommand\onedot{\futurelet\@let@token\@onedot}
\newcommand{\figref}[1]{Fig\onedot~\ref{#1}}

\newcommand{\secref}[1]{Sec\onedot~\ref{#1}}
\newcommand{\tabref}[1]{Tab\onedot~\ref{#1}}

\def\eg{\emph{e.g}\onedot} 
\def\ie{\emph{i.e}\onedot} 
 
\def\etc{\emph{etc}\onedot} \def\vs{\emph{vs}\onedot}
\def\wrt{w.r.t\onedot} 
\def\etal{\emph{et al}\onedot}

\usepackage[pagebackref=true,breaklinks=true,letterpaper=true,colorlinks,bookmarks=false]{hyperref}

\cvprfinalcopy 

\def\cvprPaperID{5401} 

\ifcvprfinal\fi

\begin{document}

\title{Factor Graph Attention\vspace{-0.5cm}}

\author{
Idan Schwartz\textsuperscript{1}, Seunghak Yu\textsuperscript{2,*}, Tamir Hazan\textsuperscript{1}, Alexander Schwing\textsuperscript{3}\\
\textsuperscript{1}Technion \hspace{1cm}\textsuperscript{2}MIT CSAIL \hspace{1cm}\textsuperscript{3}UIUC\\
{\tt\footnotesize idansc@cs.technion.ac.il, seunghak@csail.mit.edu, tamir.hazan@technion.ac.il, aschwing@illinois.edu}
\vspace{-0.5cm}
}


\maketitle

\begin{abstract}
	Dialog is an effective way to exchange information, but subtle details and nuances are extremely important. While significant progress has paved a path   to address visual dialog with algorithms,  details and nuances remain a challenge. Attention mechanisms have demonstrated compelling results to extract details in visual question answering and also provide a convincing framework for visual dialog due to their interpretability and effectiveness. However, the many data utilities that accompany visual dialog challenge  existing attention techniques. We address this issue and develop a general attention mechanism for visual dialog which operates on any number of data utilities. To this end, we design a factor graph based attention mechanism which combines any number of utility representations. We illustrate the applicability of the proposed approach on the challenging and recently introduced VisDial datasets, outperforming recent state-of-the-art methods by 1.1\% for VisDial0.9 and by 2\% for VisDial1.0 on MRR. Our ensemble model improved the MRR score on VisDial1.0 by more than 6\%. 
\end{abstract}
\blfootnote{\textsuperscript{*}Work conducted while the author was at Samsung Research.}
\section{Introduction}
Dialog is an effective way for humans to exchange information. 
 Due to this effectiveness it is an important research goal to develop artificial intelligence based agents for human-computer conversation. However, when humans talk to each other, subtle details and nuances are often very important. This importance of subtle details and nuances makes development of agents for visual dialog  a challenging endeavor. 

Recent efforts to facilitate human-computer conversation about images focus on image captioning, visual question answering, visual question generation and very recently also visual dialog. To this end, Das~\etal~\cite{visdial} collected, curated and provided to the general public an impressive dataset, which allows to design virtual assistants that can converse. Different from image captioning datasets, such as MSCOCO~\cite{lin2014microsoft}, or visual question answering datasets, such as VQA~\cite{AnatolICCV2015}, the visual dialog dataset contains short dialogs about a scene between two people. To direct the dialog, the dataset was collected by showing a caption to the first person (`questioner') which attempts to inquire more about the hidden image. The second person (`answerer') could see both the image and its caption to provide answers to these questions.
Beyond releasing the Visual Dialog dataset, to ensure a fair comparison, Das \etal~\cite{visdial} propose a particular task that can be evaluated precisely. It asks the AI system to predict the next answer given the image, the question, and a history of question-answer pairs. A variety of discriminative and generative techniques have been discussed, ranging from deep nets with Long-Short-Term-Memory (LSTM) units~\cite{HochreiterNC1997} to more involved ones with memory nets~\cite{weston2014memory} and hierarchical LSTM architectures~\cite{serban2017hierarchical}. 

\begin{figure}

\centering
\includegraphics[width=1\linewidth]{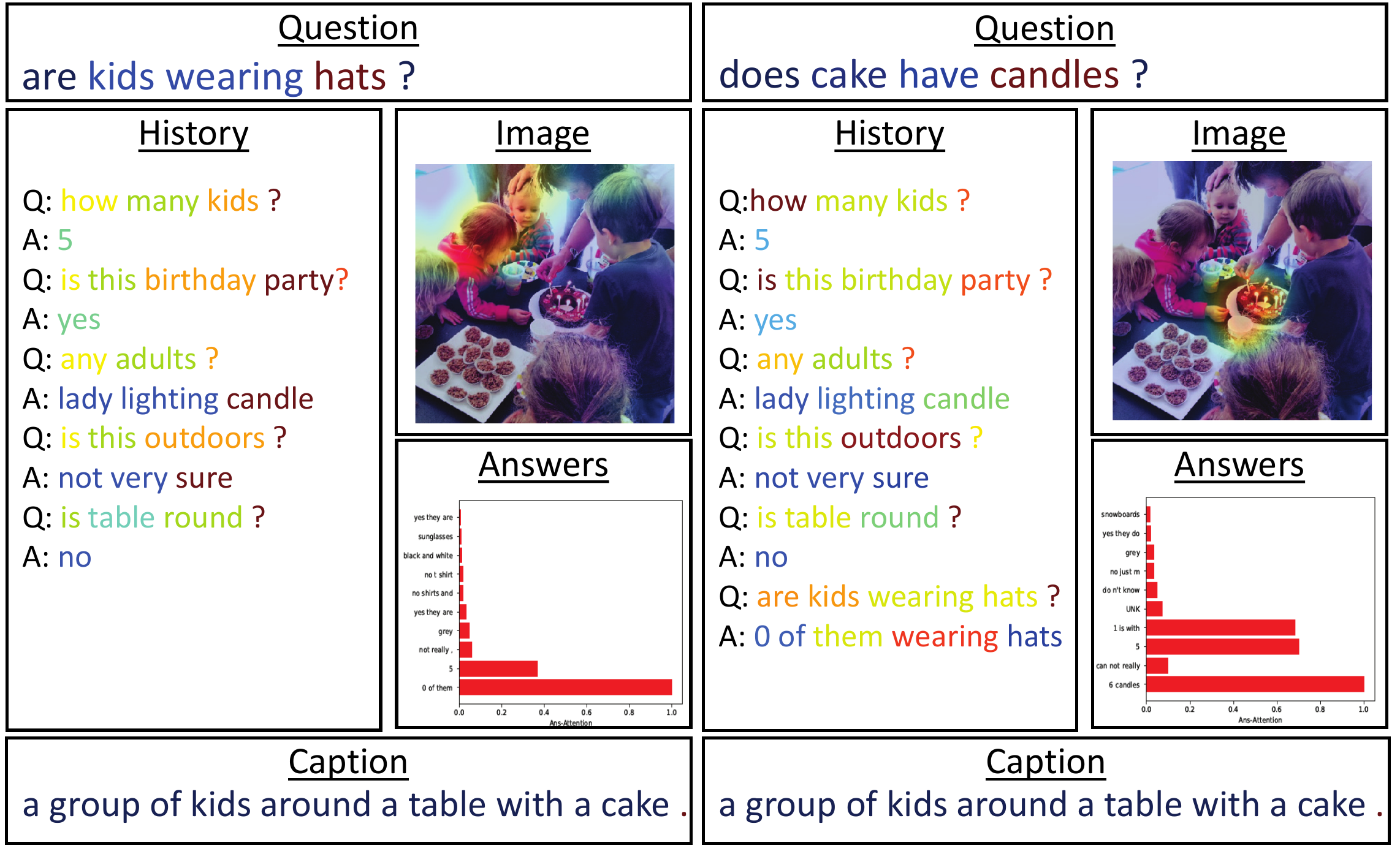}
\vspace{-0.7cm}
\caption{{\small Illustration of our factor graph attention. We show two consecutive questions in a dialog. The image attention correlates well with the question. Attention over history interactions allows our model to attend to subtle nuances. The caption focuses on the last word due to given potential priors. Attention over the answers focuses on specific  options. The attended options usually correlate with the correct answer. Note:  for readability, we chose to display only the top-10 answers out of 100 possible ones.}}
\label{fig:intro}
\vspace{-0.5cm}
\end{figure}

One of the  successful techniques to improve visual question answering is the attention  mechanism~\cite{lu2016hierarchical}.                                                                                  Due to the similarity of visual question answering and visual dialog, we envision similar improvements to be realizable. In fact, some approaches point in this direction and use a subset of the available data utilities to direct question answering~\cite{lu2017best}. 
However, in visual dialog many more ``data parts,'' \ie, the image, the question, the history and the caption are involved and have been referred to as `modalities.' To avoid confusion with the original convention/sense of the word modality, we coin the term ``utilities'' to refer to different parts of the available data.   Taking all utilities  into account makes it computationally and conceptually much more challenging to develop an effective attention mechanism. While ignoring utilities when computing attention is always an option, we argue that subtle details and nuances can only be captured adequately if we focus on all available signals. 

To address this issue we develop a general factor graph based attention mechanism which combines representations of any number of utilities. Inspired by graphical models, we use a graph based formulation to represent the attention framework, where nodes correspond to utilities  and factors model their interactions. A message passing like procedure aggregates information from modalities which are connected by edges in the graph. 

We demonstrate the efficacy of the proposed multi-utility attention mechanism on the challenging and recently introduced Visual Dialog dataset, realizing improvements up to 1.1\% on MRR. 
Moreover, we examine our model behavior using question generation proposed by~\cite{jain2018two}.    Examples of the computed attention for visual question answering are illustrated in \figref{fig:intro}.

\section{Related Work}
In recent years various machine learning techniques were developed to tackle cognitive-like multimodal tasks, which involve both vision and language processing. Image captioning~\cite{MaoARXIV2014,show_attend_tell,KarpathyCVPR2015,WangNIPS2017,ChatterjeeECCV2018,AnejaCVPR2018,DeshpandeARXIV2018} was an instrumental language+vision task, followed by visual question answering~\cite{lu2016hierarchical, SchwartzNIPS2017, kim2016hadamard,MalinowskiNIPS2014, RenNIPS2015,AnatolICCV2015,GaoNIPS2015,ZhuCVPR2016, JohnsonCVPR2017Clevr,AndreasCVPR2016,DasARXIV2016,FukuiARXIV2016,ShihCVPR2016,XuARXIV2015,SchwartzNIPS2017,XiongICML2016,NarasimhanNIPS2018,NarasimhanECCV2018,SchwartzCVPR2019} and visual question generation~\cite{RenNIPS2015,VQG,JainCVPR2017,VijayakumarARXIV2016,LiARXIV2017DualVQAVQG,BenyounesICCV2017Mutan}. 

Instrumental to cognitive tasks are attention models, that enable interpretation of the machine's cognition and often improve  performance.  
While attention mechanisms have been applied to visual question answering~\cite{FukuiARXIV2016,lu2016hierarchical, SchwartzNIPS2017,kim2017structured,XuARXIV2015,zhu2017structured}, few works have addressed visual dialog because of the many different data utilities. 
Here, we develop an attention mechanism for visual dialog, a cognitive task that was created to imitate human-like decisions~\cite{visdial}. 
We  build a general attention mechanism that is capable of capturing details. 
In the following we briefly review visual question answering and visual dialog, focusing on  the use of attention. 

\noindent\textbf{Visual Question Answering (VQA):}
Visual question answering is considered a simplified version of visual dialog since it consists of a single interaction with a given image. Some discriminative approaches include a pre-trained convolutional neural network with question embedding to predict the correct answer~\cite{Simonyan14c, MalinowskiICCV2015}. Quickly, attention mechanisms have emerged as a tool to augment the spatial attention of the image. Yang \etal~\cite{yang2016stacked} created a multi-step reasoning system via an attention model. Fukui \etal~\cite{FukuiARXIV2016} and Kim \etal~\cite{kim2016hadamard} suggested an efficient multi-modal pooling method before applying attention using a compact outer product which was later improved using the Hadamard product. Zhu \etal~\cite{zhu2017structured} treated image attention as a structured prediction task over regions, by first generating attention beliefs via unary and pairwise potentials, for which a probability distribution is inferred via loopy belief propagation.  

Alternatively, Lu \etal~\cite{lu2016hierarchical} suggested to produce Co-Attention for the image and question separately, using a hierarchical formulation. Schwartz \etal~\cite{SchwartzNIPS2017} later extended this approach for the multiple-choice VQA variant,  applying attention over image, question and answer via unary, pairwise and ternary potentials. 

\noindent\textbf{Visual Dialog:} 
D. Geman \etal~\cite{geman2015visual} were among the first to generate dialogs over images. These early attempts used only street scene images, and also restricted the conversation to templated, binary questions. 
A discriminative and generative approach was later introduced by Das \etal~\cite{visdial}, along with the largest visual dialog dataset, VisDial.  Concurrently, GuessWhat, another visual dialog dataset was published~\cite{de2017guesswhat}.  GuessWhat is a goal driven dialog dataset for object identification, while VisDial focuses on  human-like interactions. For instance, in \figref{fig:intro}, the answer for the question ``are kids wearing hats?'' is ``0 of them wearing hats,'' while a goal-driven interaction will answer with a simple ``no.'' While both types of dialogs are challenging, VisDial interactions typically consider more subtle nuances. 
Another work by Mostafazadeh \etal~\cite{mostafazadeh2017image}, focuses on conversation generation around images, instead of the content visible in images. 

The VisDial dataset is accompanied with three baselines. A vanilla approach which encodes the image, dialog and history separately and combines them subsequently (\ie, late fusion). A more complex approach based on a memory network~\cite{weston2014memory}, which maintains previous question and answer as facts in a memory bank, and learns to retrieve the appropriate fact. Lastly, a hierarchical encoding approach to capture the history~\cite{serban2017hierarchical}.  Seo \etal~\cite{seo2017visual} propose a memory network based on attention, which also addressed co-referential issues. Later, Lu \etal~\cite{lu2017best, lu2016hierarchical} combined a generative and discriminative model to choose generated answers, and also proposed history attention conditioned on the image using   hierarchical co-attention developed for visual question answering. 
Wu \etal~\cite{wu2017you}   apply attention over image, question and history representation using a Generative Adversarial Network (GAN)  to create a more human-like response. 
Jain \etal~\cite{jain2018two} developed a discriminative model that produces a binary score for each possible answer by concatenating representations of all utilities. While Jain \etal~\cite{jain2018two} also consider all utilities for  interaction prediction, our work differs in  important aspects: (1) we develop an attention mechanism that weights different representations; (2) when predicting an answer, we take information from other possible answers into account.  Recently, Kottur \etal~\cite{kottur2018visual} focused on visual co-reference resolution for visual dialog. Their approach relies on a weak supervision of a parser for reasoning~\cite{hu2017learning}, and a co-referential solver~\cite{clark2016deep}. While co-reference resolution is not the focus of our work, we found our attention model to exhibit some  co-reference resolution abilities. 

Among all attention-based techniques for Visual Dialog, the most relevant to our approach is work by Wu \etal~\cite{wu2017you} and Lu \etal~\cite{lu2017best}. Both generate  Co-Attention over the image, the question and the history representation in a hierarchical fashion. Their hierarchical approach is based on a sequential process, computing attention for one utility first and using the obtained result to generate attention for another utility subsequently. As the ordering is important, their framework is not straightforward to extend to a general multi-utility setting. 

In contrast, we develop a general attention model for any number of utilities. In the visual dialog setting, those utilities are the question in the history (10 utilities), each answer in the history (10 utilities), the caption (1 utility), the image (1 utility) and the answer representation (1 utility).  To work with a total of 23 utilities, we constructed a general attention framework that may be applied to any high-order utility setting. With our general purpose attention model we improve  results and achieve state-of-the-art performance.

To demonstrate the generality of the  approach, we also follow Jain \etal~\cite{jain2018two} and evaluate the proposed approach on choosing an appropriate question given the previous question and answer. There too we obtain state-of-the-art results. 

\noindent\textbf{Attention in General:} 
More generally, attention models have been applied to graphical data structures. For example, Graph Attention Networks use an MRF approach to embed graph-structured data, \eg, protein-protein interactions~\cite{velivckovic2017graph}. Also, attention for non-structured tasks (\eg, chain, tree) were discussed in the past~\cite{kim2017structured}. These works differ from ours in important aspects: they are used to embed a structure based model, \eg,  a graph, and 
provide  a probability distribution  
across nodes of the graph. Instead, our model 
provides attention for entities within 
each node of the graph, \eg, the words of a question or the pixels in an image.  

\section{Factor Graph Attention}
\label{sec:FGA}
In the following we describe a general framework to construct a multi-utility attention model using factor graphs. 

The factor graph is defined over \textit{utilities}, which, in the visual dialog setting, consists of an image $I$, an answer $A$, a caption $C$, and a history of past interactions $\big( H_{Q_t}, H_{A_t} \big)_{t\in\{1, \ldots, T\}}$. We subsume all utilities within the set $\mathcal{U} = \{I,A,C,\big( H_{Q_t}, H_{A_t}\big)_{t \in \{1, \ldots, T\}}\}$. In our work we have $23$ utilities (10 history questions, 10 history answers, the image, answer and caption). For notational convenience and to demonstrate the generality of the formulation we also refer to the set of utilities via  ${\cal U} = \{U_1,\ldots,U_{|{\cal U}|}\}$. Each utility $U_i \in {\mathcal{U}}$, for $i\in\{1, \ldots, |{\cal U}|\}$ consists of basic entities, \eg, a question is composed of a sequence of words and an image is composed  of spatially ordered regions. 

Formally, the $i$-th utility  $U_i$ is a $d_i \times n_i$ matrix which consists of $n_i$ entities $\hat u_i \in U_{i}$, which are the $d_i$-dimensional columns of the matrix. Each vector $\hat u_i\in U_i$ is embedded in its respective Euclidean space, \ie, $\hat u_i \in \mathbb{R}^{d_i}$, where $d_i$ is the embedding dimension of the $i$-th utility.  We use the index $u_i\in\{1, \ldots, n_i\}$ to refer to a specific column inside the matrix $U_i$, \ie, we extract the $u_i$-th column via $\hat u_i = U_{i, u_i}$. 

The $|{\cal U}|$ nodes in the factor graph each represent attention distributions over their $n_i$ utility  elements, which we call beliefs. To infer the probability we take into account two types of factors: 1) Local factors which capture information within a utility, such as their entity representation and their local interactions. 2) Joint factors which capture interactions of any subset of utilities. Due to the high number of utilities,  in our attention model, we limit ourselves to pairwise factors. Next we will explain our construction of local factors and joint factors. Note, bias terms are omitted for readability. 



\subsection{Local factors} 
\label{sec:local}
The local factors capture the local information in an employed utility $U_i$. Each utility contains entities, \ie, words in a sentence or regions in an image. There are two types of information within a utility $U_i$: \textit{Entity information}, which is extracted from an entity's vector representation $\hat u_i\in U_i$ and \textit{Entity interactions}, which  capture dependencies between two entities, such as two words in the same question or two regions in the same image. 

\noindent\textbf{Entity information:} 
This representation is obtained as the result of an embedding model, such as a Long-Short-Term-Memory (LSTM) net for sentences or a convolutional layer for image regions. Each vector representation $\hat u_i \in U_i $ has the potential to focus the model's attention to the entity the vector is representing. The potential function $\psi_i(u_i)$ is parametrized by the $i$-th utility's parameters ${V}_i$ and ${v}_i$, and is obtained via 
\beas
\psi_i(u_i)= v_i^\top  \operatorname{relu}(V_i \hat u_i).
\eeas
Hereby, $v_i \in \mathbb{R}^{d_i}, V_i \in\mathbb{R}^{d_i \times d_i}$ are trainable parameters. Recall that the index $u_i\in\{1, \ldots, n_i\}$ refers to  a specific entity. During training we also apply a dropout operation after the first linear embedding (\ie, 
$ V_i \hat u_i $). 

\noindent\textbf{Entity interactions:} The factor dependency between two elements is extracted from their vector representation. Given two indices $u_i^1, u_i^2 \in \{1, \ldots, n_i\}$, we embed the two corresponding entity representation vectors $\hat u_i^1, \hat u_i^2$ in the same Euclidean space, and  compute the factor dependency on both entities using the  dot product operation, \ie, 
\beas
\psi_{ii}(u_i^1, u_i^2)=  \left(\frac{L_{i} \hat u_i^1}{\norm{L_{i} \hat u_i^1}}\right)^\top  \left(\frac{R_{i} \hat u_i^2}{\norm{R_{i} \hat u_i^2}}\right),
\eeas
where $L_i \in \mathbb{R}^{d_i\times d_i}$, $R_{i} \in \mathbb{R}^{d_i \times d_i}$ are trainable parameters, governing the left and right arguments respectively. 


\subsection{Joint factors}
\label{sec:joint}
Joint factors capture interactions between two elements of different utilities, \eg, between a word in the question and a region in the image.  Similarly to entity interaction factors within a utility, we use
\beas
\psi_{ij}(u_i,u_j)= \left(\frac{L_{ij} \hat u_i}{\norm{L_{ij} \hat u_i}}\right)^\top  \left(\frac{R_{ji} \hat u_j}{\norm{R_{ji} \hat u_j}}\right),
\eeas
where $L_{ij} \in \mathbb{R}^{d_i \times d}$, $R_{ji} \in \mathbb{R}^{d_j \times d}$ are trainable parameters. For simplicity we let $d = \max\{d_i, d_j\}$  be the maximum dimension between the two utilities. 

To avoid a situation where pairwise scores (\eg, image and question) negatively bias another one (\eg, image and caption), proper normalization is necessary. Since the pairwise interaction scores  are generated during training, we chose a batch normalization~\cite{IoffeICML2015BatchNormalization} operation  which fixes the bias during training. 
Additionally, we applied an $L_2$ normalization on $u_i$ and $u_j$ to be of unit norm before the multiplication, \ie, we use the cosine similarity. 

\subsection{Attention, messages and beliefs}
\label{inference}
For each utility $U_i$  we infer the amount of attention that should be given to each of its elements $\hat u_i \in U_i$. Motivated by classical message-passing algorithms, we first collect all dependencies of a given utility element via
\beas 
\mu_{j \rightarrow i}(u_i) = \sum_{u_j \in \{1, \ldots, n_j\}} W_{ij}(u_i,u_j) \psi_{ij}(u_i,u_j),
\eeas
where  $ W_{ij}(u_i,u_j)  \in \mathbb{R} $ is a trainable parameter.  
We aggregate these messages from all pairwise factor dependencies and send them to a utility, in order to infer its attention belief. The inferred attention belief 
\beas
b_i(u_i) \propto \exp\left(\hat w_i p_i(u_i) +w_i \psi_i(u_i) + \sum_{j=1}^{|{\cal U}|} w_{ij} \mu_{j \rightarrow i}(u_i) \right)\!\!, 
\eeas
also uses local entity information.

\begin{figure}[t]
	\includegraphics[width=0.8\linewidth]{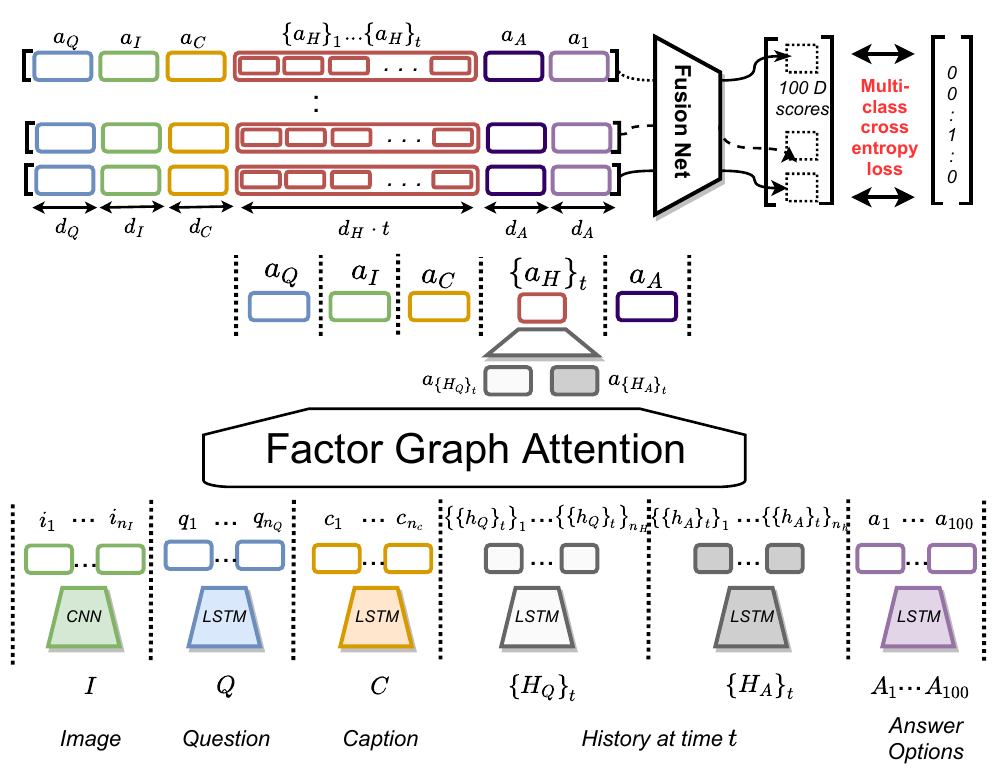}
	\put(-15,35){$\left.\begin{array}{c}~\\~\\~\end{array}\right\}$  {\footnotesize(\secref{sec:modals})}}
	\put(-15,75){$\left.\begin{array}{c}~\\~\\~\\\end{array}\right\}$  {\footnotesize(\secref{sec:atten})}}
	\put(-15,120){$\left.\begin{array}{c}~\\~\\~\\\end{array}\right\}$  {\footnotesize(\secref{sec:fusion})}}
	\vspace{-0.4cm}
	\caption{Our state-of-the-art architecture for the Visual Dialog task. Implementations details can be found in \secref{sec:visdial}. 
	}
	\label{fig:model}
	\vspace{-0.5cm}
\end{figure}

Hereby $w_{ij}, w_i$ are scalar weights learned per utility. These scalars reflect the importance of one utility with respect to the others.  For instance,  for the image belief, we find by examining these weights that the question utility is more important  than the caption utility. 
This makes sense since we want to look at relevant places for the question. 
%
Moreover, $p_i$ is a prior potential for the $i$-th utility, and $\hat w_{i}$ is a trainable parameter to calibrate the prior potential's importance. For instance, the question utility prior encourages focus of its   attention onto the last word in the question, a common practice in LSTM networks.  Using priors, we are able to steer the desired belief for a utility, while still allowing guidance of other utilities via pairwise interactions. We also experimented with priors that are updated after we infer the attention through steps, but we didn't find it to improve the results in our setup.  




Once the attention belief $b_i(u_i)$  is computed for each entity representation $\hat u_i\in U_i$, we obtain the attended vector of this utility as the average representation. This reduces the utility representation to a single vector, which is dependent on the other utilities via the belief $b_i(u_i)$: 
 \beas
 a_i = \sum_{u_i \in \{1, \ldots, n_i\}} b_i(u_i)\cdot \hat u_i.
 \eeas
Note that $a_i$ is the attended representation of utility $U_i$.

\section{Visual Dialog}
\label{sec:visdial}
We use visual dialog to demonstrate the generality of the discussed attention mechanism because many utilities are available. A general overview of the approach is illustrated in \figref{fig:model}. 
We  detail next how the general factor graph attention model is applied to visual dialog by describing (1) the utility embeddings, (2) the attention module, and (3) the fusion of attended representations for prediction.  

\begin{figure}[t]

\centering
\includegraphics[width=1\linewidth]{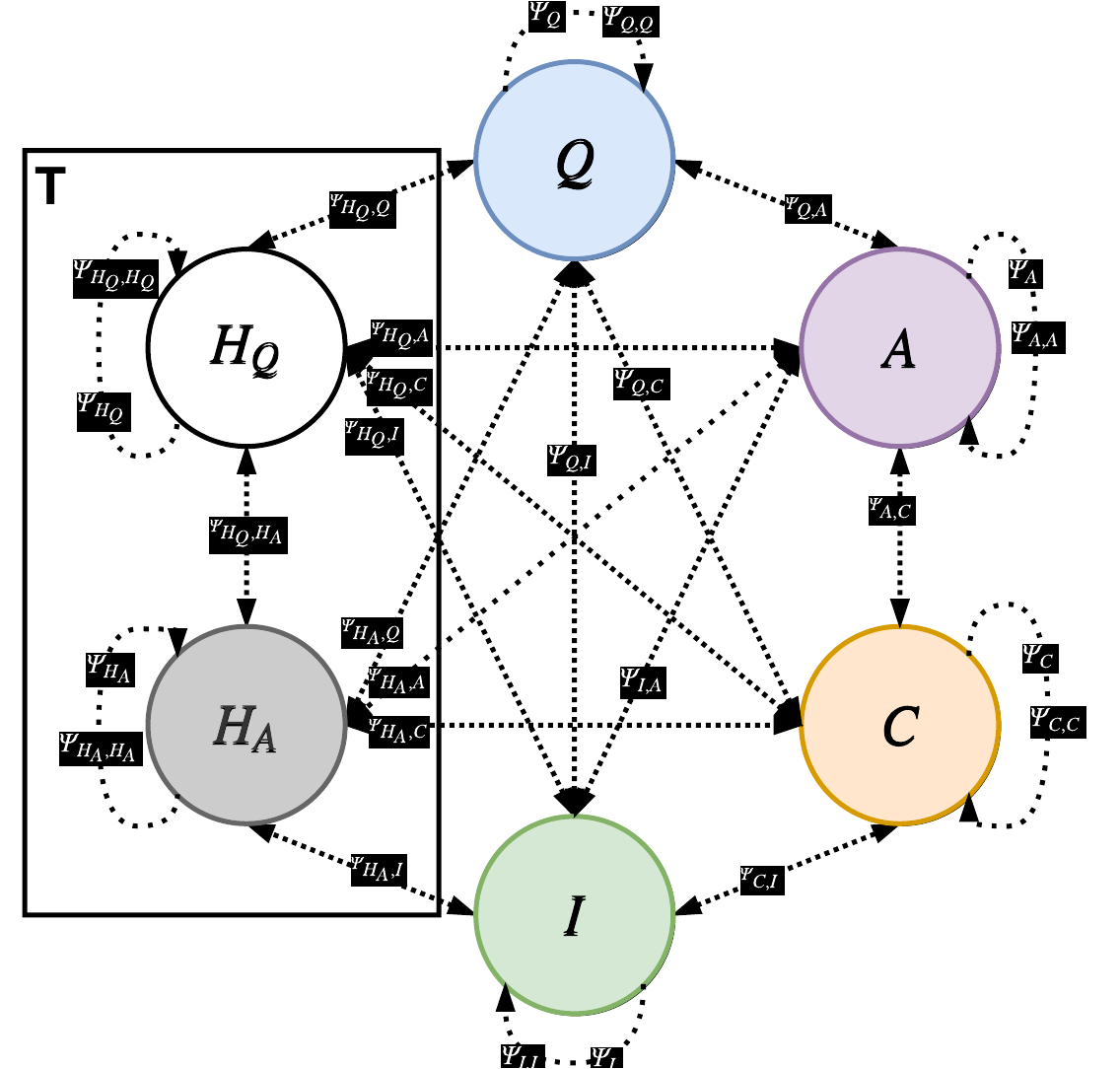}
\vspace{-0.7cm}
\caption[]{{\small A graphical representation of our attention unit. Each node represents an attention probability over the utilities' entities. To infer the probability we aggregate two types of messages: 1) A joint factor message, constructed from interactions of entities from different utilities, \eg, $ \Psi_{Q, I} $. 2) A local factor: learned from the entity representation, \eg, $ \Psi_Q $, and the self entity interactions, \eg, $ \Psi_{Q, Q} $. $ T $ is the number of history dialog interactions. }}
\label{fig:factors}
\vspace{-0.5cm}
\end{figure}

	\subsection{Utilities and Embeddings}
	\label{sec:modals}
	
	 In the following, we describe the embeddings of the image and textual utilities. 
	
	\noindent\textbf{Image utility:} 
	To represent the image regions, we use a  conv net,  pre-trained on ImageNet~\cite{imagenet_cvpr09}. Taking the output of the last convolutional layer we obtain a representation  of $7 \times 7 \times 512$. Specifically, $7\times 7$ 
	is the spatial dimension of the convolutional layer and $512$ is the number of channels/features of the representation. Following our notation in  \secref{sec:FGA}, the visual utility $U_i$ has  dimensions $n_i = 49$ and $d_i = 512$. To fine-tune this representation to our task, we feed it into another convolutional layer, with a $1\times 1$ kernel, followed by a ReLU activation and a dropout.  
	
	\noindent\textbf{Textual utilities:} 
	Our textual utilities are the caption, the question, the possible answers and the history interactions. For each textual utility $ U_i $ we embed up to $n_i$ words. Sentences with a shorter length are zero padded, while sentences of longer length are truncated. The embedding starts with a one-hot encoding representation of the word index, followed by a linear transformation. The linear transformation embeds the word index into the Euclidean space. This embedding is identical for all  textual utilities. Intuitively, usage of the same embedding ensures a better consistency between the textual utilities and we also found it to improve the results. 
	
	Each  embedded representation for each textual utility is fed into an LSTM layer, 
	which yields a representation with the appropriate embedding dimension. The caption utility $C$ and the question utility $Q$ are generated by applying a dedicated LSTM 
	on the respective embedded representation. In contrast, we embed all history questions $\big( H_{Q_t} \big)_{t\in\{1, \ldots, T\}}$ using the same LSTM model. We also embed all history answers $\big( H_{A_t} \big)_{t\in\{1, \ldots, T\}}$ using another LSTM model. 
	
	The answer utility subsumes $ n_A $ possible answers and it consists of the final decision of the model in our visual dialog system. Our answer utility uses the same LSTM to embed each of the $n_A = 100$ answers separately, the embedding of each possible answer is the LSTM hidden state of the last word in the answer.

	

	

\subsection{Attention module}
\label{sec:atten}

The attention step infers the importance of each entity in each utility, using our Factor Graph Attention (see \secref{sec:FGA}), and creates an attended representation. In the visual dialog setting, for each answer generation step we use an image $I$, a question $Q$, an answer $A$, a caption $C$, and a history of past interactions $\big( H_{Q_t}, H_{A_t} \big)_{t\in\{1, \ldots, T\}}$ (see  \figref{fig:factors} for an illustration). In the following we describe the special treatment of the different entities as well as their respective priors.


\noindent\textbf{Group utilities and dependency-relaxation:} Our factor graph attention model may have a large number of trainable parameters, as it grows quadratically with the number of utilities. To address this concern, we observe that we can group some utilities, \eg, the history answers $\big( H_{A_t} \big)_{t\in\{1, \ldots, T\}}$, and the history questions $\big( H_{Q_t} \big)_{t\in\{1, \ldots, T\}}$. To take advantage of the dependency between the group of utilities, we share the factor weights across all the group utilities. For example, for two utilities $U_{i_1}, U_{i_2} \in H_{A_t}$ we enforce the parameter sharing $v_{i_1} = v_{i_2}$, $V_{i_1} = V_{i_2}$, $L_{i_1} = L_{i_2}$, $R_{i_1} = R_{i_2}$, $L_{i_1,j} = L_{i_2,j}$ and $R_{j,i_1} = R_{j,i_2}$. Not only did it contribute to a reduced memory consumption, but we  also observed this grouping to improve the results. We  attribute the improvement to better generalization of the factors. 

The answer utility $U_i$ encodes each of the possible $n_i$ answers in a $d_i$-dimensional vector, using the LSTM hidden state at the last word. \figref{fig:intro} shows that the attention beliefs correlate with the correct answer. Note that we didn't attend separately to each possible answer. 
Doing so 
would have resulted in increased computational demand and we didn't find improved  model performance. We conjecture that due the fact that the number of words within an answer is usually small, a complete attention model on each and every word of the answer does not seem to be necessary. 


\noindent\textbf{Priors:} The prior potentials for the question and caption utilities are important in practice. For both utilities we set the prior to emphasize the last word by focusing the energy onto the last hidden state index.  We use a one hot vector with the high bit set for the last hidden state index. 




\subsection{Fusion Step}
\label{sec:fusion}


The fusion step, outlined in \figref{fig:model} combines the attended representations $a_i$ from all  utilities $\{I,A,C,\big( H_{Q_t}, H_{A_t}\big)_{t\in\{1, \ldots, T\}}\}$ to find the best answer. This is performed by creating a probability distribution $ p( u_A | I,Q,C,A,H) $ for each answer index $u_A \in\{ 1,\ldots,n_A\}$, where $n_A = 100$ is the number of possible answers. 

We denote by $a_I \in\mathbb{R}^{d_I}$ the attended image vector, $a_A  \in\mathbb{R}^{d_A}$ the attended answer vector, and $a_C \in  \mathbb{R}^{d_C} $ the attended caption vector. We construct the attended history vector $a_H \in \mathbb{R}^{d_I}$ from the attended history utilities $\big( H_{Q_t}, H_{A_t}\big)_{t\in\{1, \ldots, T\}}$. For this purpose, we start by concatenating the attended vector of each history question $a_{Q_t}$ with the concurrent history answer $a_{A_t}$, and fuse them using a linear transformation with a bias term to obtain $a_t$, which is a $ d_t $-dimensional vector. We then concatenate the attended history vectors $a_t$ for the entire dialog history  $t\in\{1,\ldots,T\}$, which results in an attended history representation $a_H \in \mathbb{R}^{d_H} $. Note that $d_H = \sum_{t=1}^T d_t$. We concatenate the image, question, caption and history  attended representations, which yields an attention representation $a \in \mathbb{R}^L $ of length $ L = d_I + d_Q + d_C + d_A + d_H$. 

Next, we combine the image, question, caption and history attended representation $a \in \mathbb{R}^L $ with the $n_A = 100$ possible answers to compute a probability for each answer. Let $U_A \in \mathbb{R}^{n_A \times d_A}$ be the answer utility, with $N = n_A = 100$ answers, while each answer is embedded in a $d_A$-dimensional space. For each answer, we denote by $\hat u_A \in \mathbb{R}^{d_A}$ 
its embedded vector. We concatenate each answer embedding with the system attention $(a,\hat u_A)$ to obtain a $(L + d_A)$-dimensional vector and feed it into a multi-layer perception  with two layers of size $ (L + d_A)/2 $ and $ (L + d_A)/4 $ respectively.  Between each layer we perform batch normalization followed by a ReLU activation. We used a dropout layer before the last fully connected layer. The obtained scores are turned into probabilities, for each answer,  using a $\operatorname{softmax}\left(\cdot\right)$ operation, which yields the posterior probability for each answer $ p( u_A | I,Q,C,A,H) $.
The approach is trained using maximum likelihood.



\begin{table}[t]
	
	\setlength{\tabcolsep}{5pt}
	\centering
	\caption{Performance of discriminative models on VisDial v0.9. Higher is better for MRR and recall@k, while lower is better for mean rank. (*) denotes use of external knowledge.}	
	\vspace{-0.3cm}
	\resizebox{\linewidth}{!}{%
		\begin{tabular}{lccccc}
			\Xhline{2\arrayrulewidth}
			\vspace{2pt}
			Model    & MRR    & R@1  & R@5   & R@10  & Mean  \\ \hline
			LF~\cite{visdial}           & 0.5807 & 43.82 & 74.68 & 84.07 & 5.78 \\
			HRE~\cite{visdial}          & 0.5846 & 44.67 & 74.50 & 84.22 & 5.72 \\
			HREA~\cite{visdial}         & 0.5868 & 44.82 & 74.81 & 84.36 & 5.66 \\
			MN~\cite{visdial}           & 0.5965 & 45.55 & 76.22 & 85.37 & 5.46 \\
			HieCoAtt-QI~\cite{lu2016hierarchical} & 0.5788 & 43.51 & 74.49 & 83.96 & 5.84 \\
			AMEM~\cite{seo2017visual}        & 0.6160 & 47.74 & 78.04 & 86.84 & 4.99 \\
			HCIAE-NP-ATT~\cite{lu2017best}        & 0.6222 & 48.48 & 78.75 & 87.59 & 4.81 \\
			SF-QIH-se-2~\cite{jain2018two}   & 0.6242 & 48.55 & 78.96 & 87.75 & 4.70 \\	
			CorefNMN~\cite{kottur2018visual}*         & 0.636 & 50.24 & 79.81& 88.51 & 4.53 \\
			CoAtt-GAN-w/ $\mathbf R_{inte}$-TF~\cite{wu2017you}         & 0.6398 & 50.29 & 80.71 & 88.81 & 4.47 \\
			CorefNMN (ResNet-152)~\cite{kottur2018visual}*         & 0.641 & 50.92 & 80.18 & 88.81 & 4.45 \\
			\hline

			FGA (VGG) & 0.6525 & 51.43 & 82.08 & 89.56 & 4.35	\\
			FGA (F-RCNNx101) & 0.6712 & 54.02 & 83.21 & 90.47 & 4.08	\\
			9$\times$FGA  (VGG) & \textbf{0.6892} & \textbf{55.16} & \textbf{86.26} & \textbf{92.95} & \textbf{3.39} \\ 		
			\Xhline{2\arrayrulewidth}
	\end{tabular}}
	\vspace{-0.3cm}
	\label{tab:baselines}
\end{table}

\begin{table}[t]

	\centering
	\caption{Performance on the question generation task. Higher is better for MRR and recall@k, while lower is better for mean rank.}
	\vspace{-0.3cm}	
	\resizebox{\linewidth}{!}{%
	\begin{tabular}{lccccc}
		\Xhline{2\arrayrulewidth}

		Model    & MRR    & R@1  & R@5   & R@10  & Mean  \\ \hline
		SF-QIH-se-2~\cite{jain2018two}   & 0.4060 & 26.76 & 55.17 & 70.39 & 9.32 \\
		\hline
		FGA & \textbf{0.4138} & \textbf{27.42} & \textbf{56.33} & \textbf{71.32} & \textbf{9.1}	\\	 
		\Xhline{2\arrayrulewidth}
	\end{tabular}}
\vspace{-0.5cm}
	\label{gen-results}
\end{table}

\section{Results}

In the following we evaluate the proposed factor graph attention (FGA) approach on the Visual dialog dataset, which we briefly describe first. Our code is publicly available\footnote{https://github.com/idansc/fga}. 

\noindent{\bf Visual Dialog Dataset:} We used VisDial v0.9 to train the model. The dataset consists  of approx.\ 120k images from COCO~\cite{lin2014microsoft}. Each image is annotated with a dialog of 10 questions and corresponding answers, for a total of approx.\ 1.2M dialog question-answer pairs. In the discriminative setup, each question-answer pair is given 100 plausible possible answers, the model needs to choose from. We follow~\cite{visdial} and split the data into 80k images for train, 40k for test  and 3k for validation. 

\noindent{\bf{Experimental setup:}} We used a batch size of 64. We set the word embedding dimension to $ d_E = 128 $, and the utility embeddings to $ d_Q =  512 $ and  $ d_C = 128 $. For each question or answer in the history we use $ d_{{H_Q}_i} = d_{{H_A}_i} = 128 $. For each possible answer we use $ d_a = 512 $. The lengths are set equally for all textual utilities $ n_Q = n_C = n_a = n_{H_Q} = n_{H_A} = 20$. The VisDial history consists of $ T = 10 $ questions with their answers. For our image representation we use  the last conv layer of VGG having  dimensions of $ 7\times 7\times 512 $. After flattening the 2D spatial dimension, $ n_I = 49 $. The dropout parameter after the image embedding is set to $0.5$, the dropout parameter before the last fc layer is set to $0.3$. 

\begin{table}[t]

	\setlength{\tabcolsep}{5pt}
	\centering
	\caption{Performance of discriminative models on VisDial v1.0 test-std. Higher is better for MRR and recall@k, while lower is better for mean rank and NDCG. (*) denotes use of external knowledge.
	}	
	\vspace{-0.3cm}
	\resizebox{\linewidth}{!}{%
	\begin{tabular}{lcccccc}
			\Xhline{2\arrayrulewidth}
			\vspace{2pt}
			Model    & MRR    & R@1  & R@5   & R@10  & Mean & NDCG  \\ \hline
			LF~\cite{visdial} & 0.554 & 40.95 & 72.45 & 82.83 & 5.95 & 0.453 \\
			HRE~\cite{visdial} & 0.542 & 39.93 & 70.45 & 81.50 & 6.41 & 0.455 \\
			MN~\cite{visdial} & 0.555 & 40.98 & 72.30 & 83.30 & 5.92 & 0.475 \\
			CorefNMN (ResNet-152)~\cite{kottur2018visual}*   & 0.615 & 47.55  & 78.10 & 88.80 & 4.40 & 0.547 \\
			NMN (ResNet-152)~\cite{hu2017learning}* & 0.588 & 44.15 & 76.88 & 86.88 & 4.81 &\textbf{ 0.581} \\
			\hline
			FGA (VGG) & 0.637 & 49.58 &80.97 & 88.55 &   4.51 & 0.521	\\	
			FGA (F-RCNNx101) & 0.662 & 52.75 & 82.92 & 91.07 & 3.8 &  0.569	\\
			5$\times$FGA (VGG) & 0.673 & 53.40 &  85.28 & 92.70  & 3.54 &  0.545 \\
			5$\times$FGA (F-RCNNx101) & \textbf{0.693} & \textbf{55.65} &  \textbf{86.73} & \textbf{94.05} & \textbf{3.14} &  0.572 	\\ 
			\Xhline{2\arrayrulewidth}

		\end{tabular}}
\vspace{-0.3cm}
		\label{tab:baselines2}
\end{table}

\begin{table}[t]

	\centering
	\caption{Attention-related ablation analysis.}	
	\vspace{-0.3cm}
	\resizebox{\linewidth}{!}{%
	\begin{tabular}{lccccc}
		\Xhline{2\arrayrulewidth}
		\vspace{2pt}
		Model    & MRR    & R@1  & R@5   & R@10  & Mean  \\ \hline
		No Attention & 0.6249 & 48.67 & 78.95 & 87.73 & 4.69	\\						
		No BatchNorm & 0.6301 & 49.23 & 79.65 & 88.32 & 4.55	\\			
		No Local-Interactions   & 0.6369 & 50.17 & 79.92 & 88.33 & 4.55 \\		
		No Local-Information   & 0.6425 & 50.12 & 81.49 & 89.34 & 4.37 \\	
		No Priors   & 0.6451 & 50.57 & 81.37 & 89.00 & 4.47 \\	
				\hline		
		FGA & \textbf{0.6525} & \textbf{51.43} & \textbf{82.08} & \textbf{89.56} & \textbf{4.35}	\\

		\Xhline{2\arrayrulewidth}
	\end{tabular}}
\vspace{-0.3cm}
	\label{tab:abl-atten}
\end{table}

\begin{table}[t]
	
	\centering
	\caption{Utility-related ablation analysis.}	
	\vspace{-0.3cm}
	\resizebox{\linewidth}{!}{%
		\begin{tabular}{lccccc}
			\Xhline{2\arrayrulewidth}
			\vspace{2pt}
			Model    & MRR    & R@1  & R@5   & R@10  & Mean  \\ \hline
			No Answer Utility & 0.6294 & 49.35 & 79.31 & 88.10 & 4.63	\\	
			No History Attention & 0.6449 & 50.74 & 81.07 & 88.86 & 4.48	\\
			Answers Fine-attention & 0.6478 & 50.80 & 81.86 & 89.25 & 4.46	\\
			History No Fine-attention  & 0.6494 & 51.17 & 81.56 & 89.13 & 4.43	\\

			\hline
			FGA & \textbf{0.6525} & \textbf{51.43} & \textbf{82.08} & \textbf{89.56} & \textbf{4.35}	\\	 
			\Xhline{2\arrayrulewidth}
	\end{tabular}}
	\vspace{-0.5cm}
	\label{tab:abl-util}
\end{table}

\begin{figure*}[t]
\vspace{-0.5cm}
\centering
\includegraphics[width=1\linewidth]{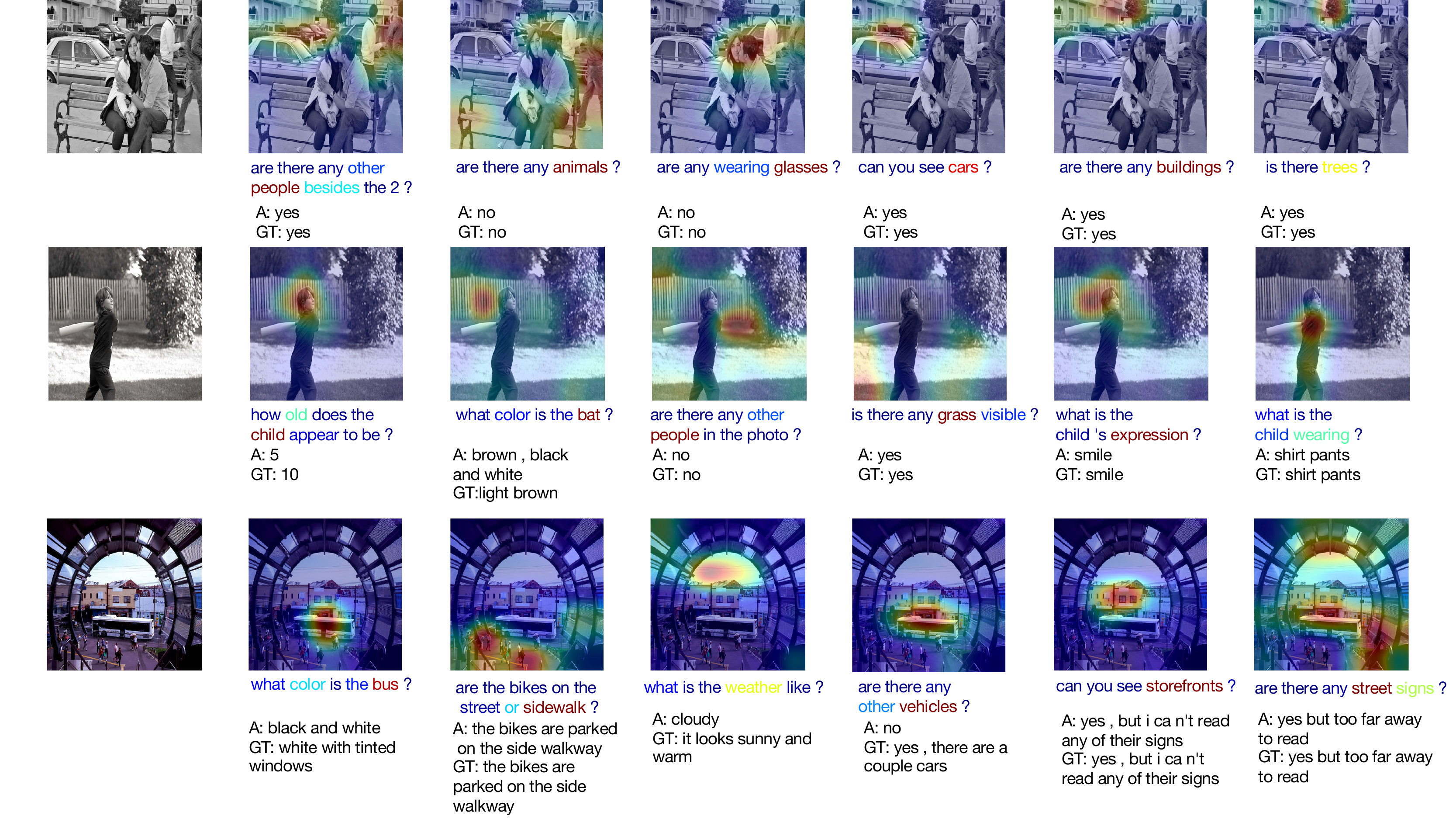}
\vspace{-0.8cm}
\caption{An illustration of question and image attention over a series of interactions for the same dialog. In addition we provide the ground truth answer, \ie, GT, and our predicted answer, \ie, A.}
	\label{fig:q_i}
	\vspace{-0.5cm}
\end{figure*}

\noindent{\bf{Training:}} The total amount of trainable parameters in our model is $17,848,416$. We initialized all the weights in the model using Kaiming normal initialization \cite{he2015delving}. To train the model we used a multi-class cross entropy loss, where each possible answer represents a class. We used Adam optimizer with a learning rate of $10^{-3}$. We evaluate our performance on the validation set after each epoch to determine when to stop our training.

\subsection{Quantitative Evaluation}
\noindent{\bf Evaluation metrics:} Evaluating dialog systems, or any other generative tasks  is challenging~\cite{LiuEMNLP2016HowNOT}. We follow~\cite{visdial}  and evaluate each individual response at each of the $ T = 10 $ rounds in a multiple-choice setup. 
The model is hence evaluated on retrieval metrics:  Recall@k is the percentage of questions where the human response was part of the top $ k $ predicted answers. Mean rank is the average rank allotted by a model to the human response, hence a lower score is desired. Mean Reciprocal Rank (MRR) is defined as $\frac{1}{|{Q}|}\sum_{i=1}^{|Q|}\frac{1}{\text{rank}_i}$, where $\text{rank}_i$ is the rank of the human response, and $ Q $ is the set of all questions. The perfect score, \ie, MRR = 1 is achieved when the  human response is consistently ranked first.

\noindent{\bf Visual question answering comparison:} We first compare against a variety of baselines (see \tabref{tab:baselines}). Note that almost all of the baselines (except LF, HRE and MN and SF-QIH-se-2) use attention, \ie, attention is an important element in any model. 
Note that our model uses the entire set of answers to predict each answer's score, \ie, we use $p(u_i|A, I, Q, C, H) $ This is in contrast to SF-QIH-se-2, which doesn't use attention and models $ p(u_i|\hat{u_i}, I, Q, C, H) $.
Notable as well, the current state-of-the-art model, CoAtt-GAN~\cite{wu2017you}, used the largest amount of utilities to attend to, \ie, image, question and history. Because CoAtt-GAN  uses a hierarchical  approach, the ability to further improve the reasoning system is  challenging and manual work.  In contrast, our general attention mechanism allows to attend to the entire set of cues  in the dataset,  letting the model automatically choose the more relevant cues.  
We refer the readers to the appendix for analysis of utility-importance via importance score. 
As can be seen from \tabref{tab:baselines}, this results in a significant improvement of performance, even when compared to the very recently published baselines~\cite{jain2018two, wu2017you, kottur2018visual}. 
We also report an ensemble of 9 models which differ only by the initial seed. 
We emphasize that our approach only uses VGG16. Lastly, some baselines report to use GloVe to initialize the word embeddings, while we didn't use any pre-trained embedding weights. 

Our attention model is very efficient to train. 
Our state-of-the-art score is achieved after only 4 epochs. Each epoch takes approximately 2 hours on a standard machine with an Nvidia Tesla M40 GPU. In contrast, CorefNMN~\cite{kottur2018visual}, has 100M parameters and takes 33 hours to train on a Titan X. Both~\cite{lu2017best,wu2017you}  report that more than 25 epochs 101M parameters and 50 hours were required for training. 

\noindent{\bf Visual question generation comparison:}  
To assess question generation,  \cite{jain2018two} proposed to predict the next question given the previous  question and answer. Their introduced question prediction dataset is based on VisDial v0.9, along with a collected set of 100 question candidates. 

 We adapted to this task, by changing the input utilities to the previous interaction $ (Q+A)_{t-1} $ instead of the current question $ Q_t $. Our model also improves  previous state-of-the-art results (see \tabref{gen-results}). 

\noindent{\bf Visual Dialog Challenge:} 
Recently, VisDial v1.0 was released as part of the  Visual Dialog challenge, where 123,287 images are used for training, 2,000 images for validation, and 8,000 images for testing.  For the test split each image consists of only 1 interaction, at some point of time in the dialog. Furthermore, an additional metric, normalized discounted cumulative gain (NDCG), was introduced. NDCG uses dense annotations, \ie, the entire set of candidate answers is annotated as true or wrong. The metric penalizes  low ranking  correct answers,  addressing issues when the set of answers contains more than one plausible result. 

Our submission to the challenge significantly improved all metrics except for NDCG. We report our results in~\tabref{tab:baselines2} on test-std, a 4,000 image split, the other 4,000 image split was preserved for the challenge. 
While the challenge did allow use of any external resources to improve the model, we only changed  our approach to use an  ensemble of 5 trained Factor Graph Attention models which were initialized randomly. All  other top teams used external data in form of detection features on top of ResNet-152, inspired by Top-Bottom attention~\cite{anderson2018bottom}. These features are  expensive to extract, and  use external detector information. 

Our model  used only the single ground truth answer to train. Therefore it is expected that our model isn't optimized \wrt the NDCG metric. However, given the small subset of densely annotated samples (2,000 out of the 123,287 train images), it is hard to carefully analyze this result. 
 

\begin{figure}
\vspace{-0.0cm}
	\centering
	\includegraphics[width=1\linewidth]{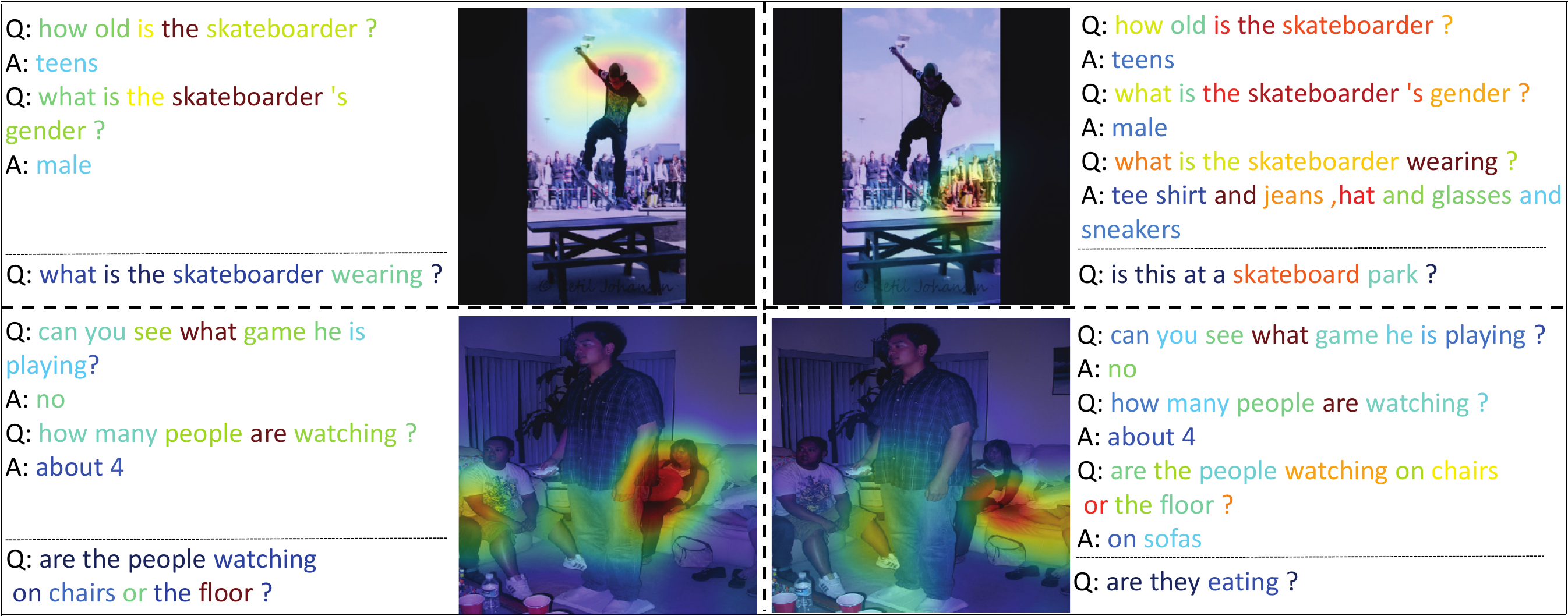}
	\vspace{-0.7cm}
	\caption{Illustration of history attention for 2 interactions. We observe small nuances of history to be useful to answer questions, and improve co-reference resolution.}
	\label{fig:coref}
	\vspace{-0.5cm}
\end{figure}

\noindent{\bf Ablation Study:} 
We asses (1) design choices of our factor graph attention; and (2) utility ablation focusing on history and answer cues as they are a unique aspect of our work. 
(1)  In \tabref{tab:abl-atten} we see that FGA improves the MRR of a model without attention by 3\% (0.6249 \vs 0.6653). This ablation study shows that attention is crucial for VisDial. Removing local-information drops MRR to 0.6425. When omitting local-interactions, \ie, a score based on interactions of embedding representations of a utility, the MRR drops to 0.6369. BatchNorm over pairwise interactions is crucial. Without BatchNorm MRR drops to 0.6301. Removing prior information, \eg, a high prior potential for the last word in the question is less crucial, dropping MRR to 0.6451. (2)  Our history attention attends separately to questions and answers in the history. In contrast, classical methods~\cite{WuARXIV2016, seo2017visual} attend over history locations only. Based on \tabref{tab:abl-util}, we note that our fine-grained history attention improves MRR from 0.6494 to 0.6525. Without the answers utility, performance on MRR drops significantly from 0.6525 to 0.6294. If we attend to each word in the answers separately, \ie, `Answers Fine-Attention,' performance drops to 0.6478.


\noindent{\bf Other Datasets:}  When we replace the attention unit of other methods with our FGA unit we observe improvements in visual question answering (VQA) and audio-visual scene aware dialog (AVSD)~\cite{AnatolICCV2015, alamri2018audio}. For VQA v1.0 we increase validation set accuracy from 57.0 to 57.3 (no tuning) by replacing the alternating and parallel attention~\cite{lu2016hierarchical}. 
For AVSD, we 
improve Hori \etal~\cite{hori2018end}  which report a CIDEr score of 0.733 to 0.806. We used FGA to attend to all video cues as well as the question. This differs from Hori \etal who mix the question representation with  video-related cues (\eg, I3D features, optical flow and audio features), and aggregate them to generate attention. Other components remain the same. Our flexible framework is instrumental for this improvement. 

\noindent{\bf Bottom-up  Features:} We follow Anderson \etal~\cite{anderson2018bottom} and use bottom-up features of 36 proposals from images. Equipped with bottom-up features as image representation our ensemble network increase MRR score on VisDial v1.0 by 2\% (0.673 vs 0.693). For a single model we observe a similar boost in performance (0.6525 vs 0.6712) on VisDial v0.9.

\subsection{Qualitative Evaluation}
Attention is an important tool not only because it boosts performance, but also because it yields a weak form of interpretability. By illustrating the attention beliefs, we can observe the reasoning process of the model. In \figref{fig:q_i} we provide  co-attention of image and question. The first row shows dialogs with yes/no questions. We observe the question attention to focus on the indicative word, \eg,  people, animals, buildings, cars, \etc, while the image attention performs  detection and attends to the relevant area of the image. 
For the second row, again we observe plausible attention behavior. An interesting failure-case: when asked about the color of the bat, the ground-truth answer was ``light brown,'' while our model answered ``brown, black and white'' instead. A possible explanation is related to the fact that the image is in black and white. 
The last line shows that question-answering type of task is always debatable. For the question ``what is the weather like?'' the model answered ``cloudy,'' while the ground truth is ``it looks sunny and warm.'' While it does look sunny, the model attends to clouds and the model answer likely isn't entirely wrong. 

\begin{figure}[t]
\vspace{-0.0cm}
\centering
\includegraphics[width=\linewidth]{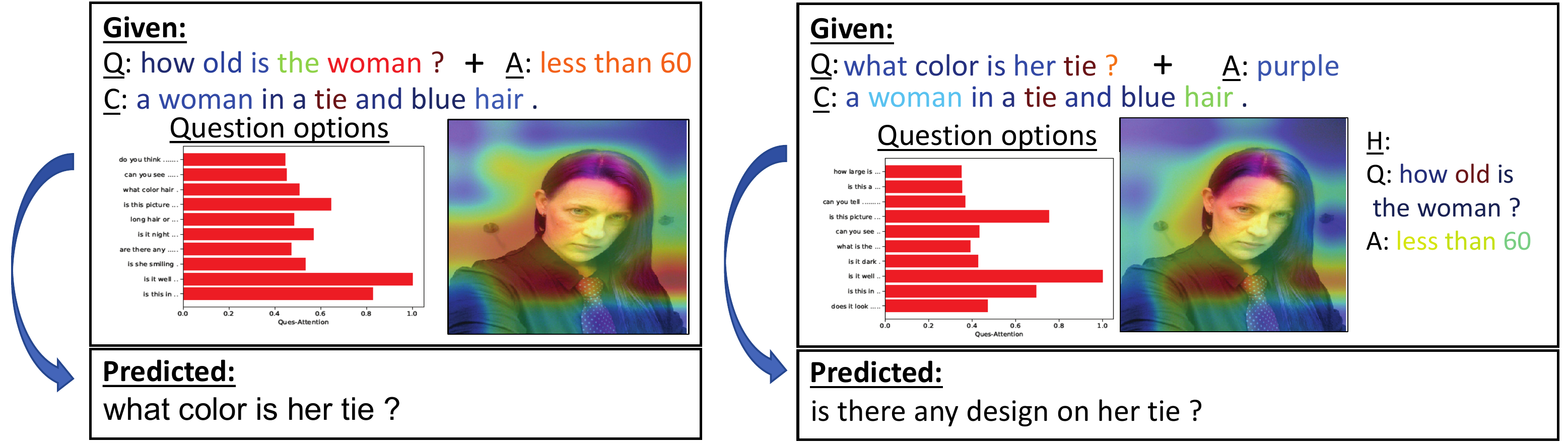}
\includegraphics[width=\linewidth]{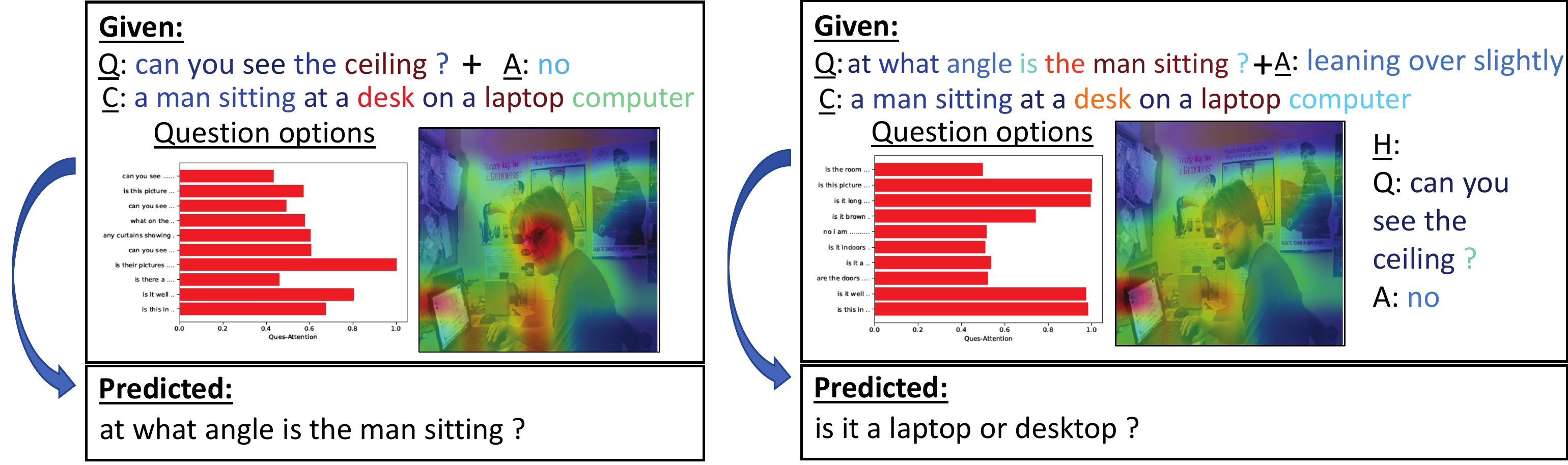}
	\vspace{-0.7cm}
\caption{Illustration of 2 step interaction using  visual question generation and illustration of the involved modalities. The classifier receives the previous question and answer, to predict a new one.}
\label{fig:qdials}
	\vspace{-0.3cm}
\end{figure}

Next, in \figref{fig:coref}, we show how attention is useful when applied over each question in the history. In the first row, for the question ``is this at a skateboarder park?'', the skateboard related terms in the history are given more weight. Another use case of  attention is co-reference resolution.  We highlight those results  in the second row: the word ``they'' in the second question refers to people in the background, which remain the focus of the attention model. 



Lastly, in \figref{fig:qdials}, we evaluate  question generation and let the model interact with the answer predictor.  We show how  complete dialogs can be generated in a discriminative manner. We first observe that attention for question generation  is noisier. This seems intuitive because asking  a question requires a broader focus than answering. 
Nonetheless, visual input is important. For the second row second image, ``at what angle is the man sitting?'' the model attends mostly to the man, and for the question ``is it a laptop or desktop?''  image attention focuses on the laptop. Also, in both cases the caption attention is useful. For instance, in the first row, the word ``tie'' is picked to generate two relevant questions. This nicely illustrates how the proposed model adapts to  tasks, when the importance of different data cues changes.





\section{Conclusion}
\label{sec:conc}
\vspace{-0.2cm}
We developed a general factor graph based attention mechanism which can operate on any number of utilities. We showed applicability of the proposed attention mechanism  on the recently introduced visual dialog dataset and outperformed existing baselines by 1.1\% on MRR.

\noindent\textbf{Acknowledgments:} This research was supported in part by The Israel Science Foundation (grant
No. 948/15), and by NSF under
Grant No.\ 1718221, Samsung, and 3M.  

{\small
	\bibliographystyle{ieee}
	\bibliography{alex}
}

\end{document}